\documentclass{article}

    \PassOptionsToPackage{numbers, compress}{natbib}

\usepackage[final]{neurips_2022}

\usepackage[utf8]{inputenc} 
\usepackage[T1]{fontenc}    
\usepackage{hyperref}       
\usepackage{url}            
\usepackage{booktabs}       
\usepackage{amsfonts}       
\usepackage{nicefrac}       
\usepackage{microtype}      
\usepackage{xcolor}         
\usepackage{amsmath}
\usepackage{graphicx}
\usepackage{wrapfig}
\usepackage{subcaption}

\usepackage{ulem}

\title{Transfer Learning with Physics-Informed Neural Networks for Efficient Simulation of Branched Flows}

\author{
    Rapha\"el Pellegrin \\
    Harvard University \\
    \texttt{raphaelpellegrin@fas.harvard.edu} \\
    \And
    Blake Bullwinkel \\
    Harvard University \\
    \texttt{jbullwinkel@fas.harvard.edu}
    \And
    Marios Mattheakis \\
    Harvard University \\
    \texttt{mariosmat@seas.harvard.edu}
    \And
    Pavlos Protopapas \\
    Harvard University \\
    \texttt{pavlos@seas.harvard.edu}
}

\begin{document}

\maketitle

\begin{abstract}
Physics-Informed Neural Networks (PINNs) offer a promising approach to solving differential equations and, more generally, to applying deep learning to problems in the physical sciences. 
We adopt a recently developed transfer learning approach for  PINNs and introduce a multi-head model to efficiently obtain accurate solutions to nonlinear systems of ordinary differential equations with random potentials.
In particular, we apply the method to simulate stochastic branched flows, a universal phenomenon in random wave dynamics.  
Finally, we compare the results achieved by feed forward and GAN-based PINNs on two physically relevant transfer learning tasks and show that our methods provide significant computational speedups  in comparison to standard PINNs trained from scratch.

\end{abstract}

\section{Introduction}

Differential equations are used to describe a plethora of phenomena in the physical sciences but most cannot be solved analytically. Traditionally, numerical methods have been used to approximate solutions to differential equations. Recently, Physics-Informed Neural Networks (PINNs) have emerged as an attractive alternative offering several compelling advantages. In particular, PINNs: provide solutions that are in closed form, offer a more accurate interpolation scheme \cite{lagaris1998neural}, are more robust to the ``curse of dimensionality'' \cite{him_dim_proof, nn_highdim_pdes, highdim_nn_pde_forward_backward, dgm_highdim_pde_nn}, and do not accumulate numerical errors \cite{jin2020unsupervised, mattheakis2020hamiltonian}. 

PINNs are typically trained to solve only a single configuration of a given system (e.g., a single initial condition or set of system parameters) at once, making their practical use computationally inefficient. 
More recently, it was shown that one-shot transfer learning can be used to obtain accurate solutions to linear systems of ordinary differential equations (ODEs) and partial differential equations (PDEs), thereby eliminating the need to train the network from scratch for a new linear system \cite{desai2021one,mattheakis2021unsupervised}.

In this work, we build upon \cite{desai2021one} by proposing a method that can be applied to {non-linear} systems. This method consists of two phases. First, we train a base neural network with multiple output heads, solving the system for a range of different configurations (e.g., initial conditions or potentials). We thus learn a representative basis that captures the underlying dynamics. Second, we freeze the weights of the base network and fine-tune new linear heads on a secondary transfer learning task. In doing so, we adapt the pre-trained base from one task to another and cut computational costs significantly. We demonstrate the efficacy of our approach using a system of non-linear ODEs that describes the trajectory of a particle through a weak random potential and can be used to model a universal phenomenon called branched flow \cite{degueldre2016random, mattheakis2018emergence}.\footnote{All code is publicly available at https://github.com/RaphaelPellegrin/Transfer-Learning-with-PINNs-for-Efficient-Simulation-of-Branched-Flows.git.} 

\section{Background}

\subsection{PINN Models}

This work uses PINNs that are trained in an unsupervised manner, as detailed below. We compare the performance achieved by feed forward neural networks (FFNN) and GAN models.

\textbf{FFNN:}  The standard unsupervised neural network approach was introduced by \citet{lagaris1998neural} and can be used to solve differential equations of the form
\begin{equation}
    \label{eq:diffeq}
    \mathcal{L}\left(u(t,\bold{x})\right)=0,
\end{equation}
where $\bold{x}=(x_1,\dots, x_n)$, $u: \mathbb{R}^{n+1} \rightarrow \mathbb{R}$ and $\mathcal{L}$ is a differential operator. During training, we sample $(t, \bold{x})$ from the domain of the equation $D$ and use this vector as input to a FFNN, which outputs the neural solution $u_{\theta}(t,\bold{x})$. We re-parametrise this output into $\tilde{u}_\theta(t, \bold{x})$ to satisfy initial and boundary conditions exactly. Using automatic differentiation, we can compute the derivatives of this output with respect to each of the independent variables and build the loss by summing the squared residual over $M$ training points,
\begin{equation}
    \label{eq:l2_norm_loss}
    \frac{1}{M}\sum_{(t,\mathbf{x})\in D} \mathcal{L}\left(\tilde{u}_{\theta}(t,\bold{x})\right)^2.
\end{equation}
Note that if the network $\tilde{u}_{\theta}$ perfectly satisfies Equation \ref{eq:diffeq}, then Equation \ref{eq:l2_norm_loss} will be zero. 

\textbf{DEQGAN:} \citet{bullwinkel2022deqgan} noted that there is no theoretical reason to use the $L_2$ norm of the residuals over any other loss function and proposed DEQGAN, which extends the FFNN method to GANs and can be thought of as ``learning the loss function.'' Rather than computing a loss over the equation residuals, DEQGAN labels these vectors ``fake'' data samples and zero-centered Gaussian noise as ``real'' data samples. As the discriminator gets better at classifying these samples, the generator $\tilde{u}_\theta$ is forced to propose solutions such that the equation residuals are increasingly indistinguishable from a vector of zeros, thereby approximating the solution to the differential equation.

\subsection{Transfer Learning with Multi-Head PINNs}

\begin{wrapfigure}{r}{0.5\textwidth}
\centering
\includegraphics[width=0.95\linewidth]{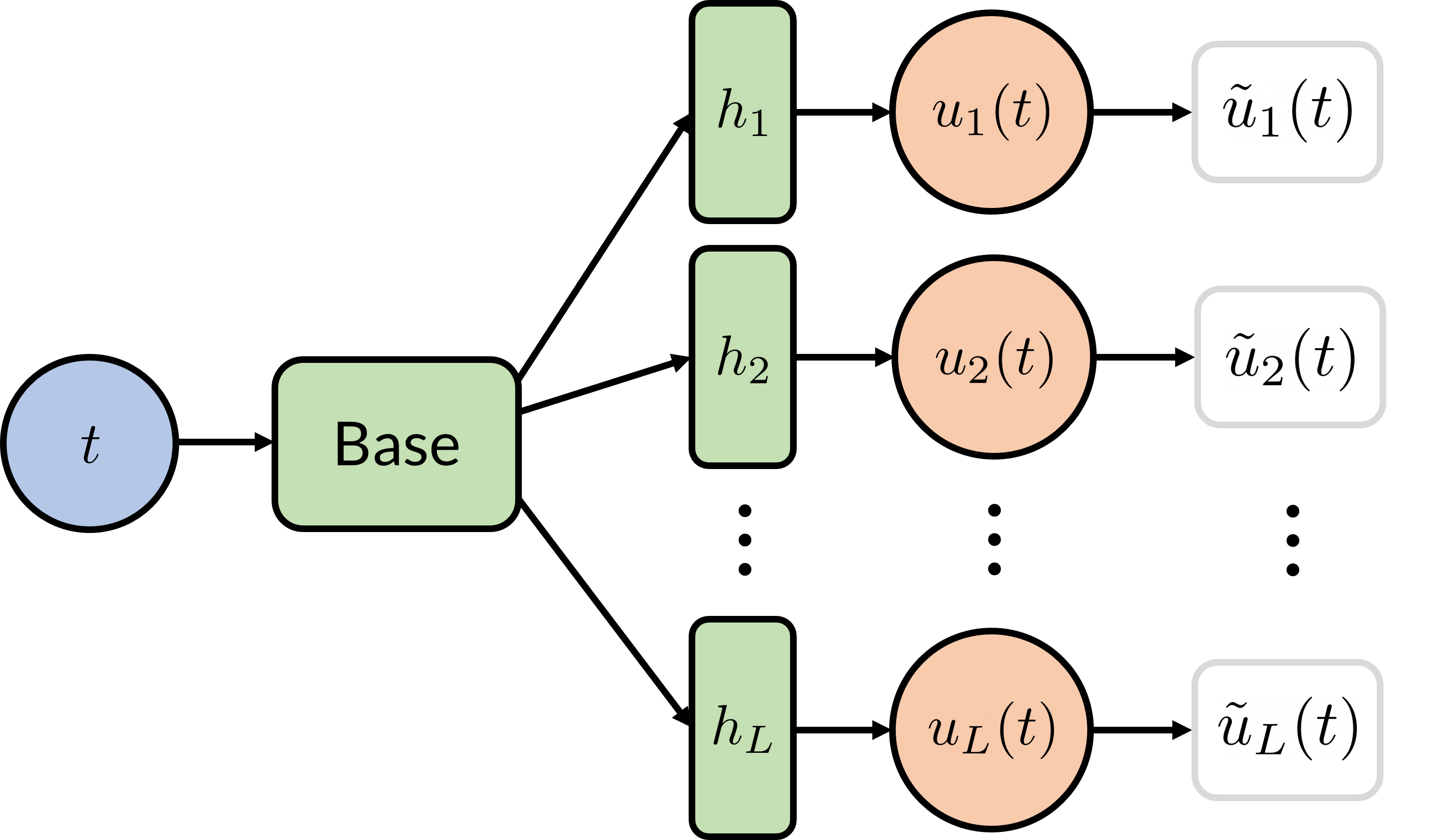}
\caption{Multi-head PINN architecture. Each output head $h_l$ is responsible for generating the solution to the $l^\text{th}$ initial condition.}
\label{fig:multi_head_architecture}
\end{wrapfigure}

Figure~\ref{fig:multi_head_architecture} illustrates the multi-head architecture that we apply to FFNN and DEQGAN models to perform transfer learning. The output of the base neural network is passed to heads $h_1,...,h_L,$ each of which corresponds to the solution to the system at a particular initial condition. Importantly, this architecture can be used to apply transfer learning to non-linear problems; in this work, we consider one such system of ODEs that is used to model particles moving through a weak random potential. 

Transfer learning with multi-head PINNs is performed in two stages. First, we train the multi-head model on a given set of initial conditions until convergence. Next, we freeze the weights of the base network and fine-tune only the output heads on a second set of initial conditions. As detailed below, we use this procedure to perform two transfer learning tasks: 1) Initial Condition Transfer, which involves fine-tuning the heads on initial conditions that were not used to train the base. 2) Potential Transfer Learning, an even more challenging task that allows us to obtain solutions for new initial conditions and a different potential than the one used to train the base. 
Our results on these tasks suggest that the base is able to learn highly general properties of the system.

\section{Experimental Results}

\subsection{Branched Flow} 
Stochastic branched flow is a universal wave phenomenon that occurs when waves propagate in  random environments. Branching has been observed in tsunami waves \cite{degueldre2016random}, electronic flows in graphene \cite{mattheakis2018emergence}, and electromagnetic waves in gravitational fields \cite{loutsenko2018role}. We can model a two dimensional branched flow by considering a particle with position $\bold{x}=(x,y)$ and velocity $\bold{p}=(p_x,p_y)$, both functions of time $t$, traveling through a weak random potential $V(x,y)$. With the Hamiltonian $H(\bold{x},\bold{p})=||\bold{p}||^2_2/2+V(\bold{x})$
we obtain Hamilton's equations, given by the following system of ODEs
\begin{equation}
\left\{ \begin{aligned} 
  \dot{x}(t) &= p_x(t)\\
  \dot{y}(t) &= p_y(t)\\
  \dot{p_x}(t) &= -\frac{\partial V(x(t),y(t))}{\partial x}\\
  \dot{p_y}(t) &= -\frac{\partial V(x(t),y(t))}{\partial y},
\end{aligned} \right.
\end{equation}
For a plane wave, the initial conditions at $t=0$ can be chosen as $(x(0),y(0),p_x(0),p_y(0))=(0,y(0),1,0)$ \cite{mattheakis2018emergence}.

We build  random potentials by summing $K$ randomly distributed Gaussian functions with covariance matrix $\sigma^2 I_2 \in \mathbb{R}^{2\times 2}$ and means $\mu_i \in \mathbb{R}^2$, $i=1,\dots, K$, and scaling the result by $-A$, where $A \in \mathbb{R}^+$, as in \cite{mattheakis2018emergence}. That is,
\begin{equation} \label{pot} V(\bold{x})=-\frac{A}{2\pi \sigma^2}\sum_{i=1}^{K} \exp\left(-\frac{1}{2\pi \sigma^2}|| \bold{x}-\bold{\mu}_i ||_2^2\right).  \end{equation}

In the experiments presented below, we use $K=10,~A=0.1, ~\sigma =0.1$.

\begin{figure}[h]
\begin{subfigure}[h]{0.5\linewidth}
\includegraphics[width=\linewidth]{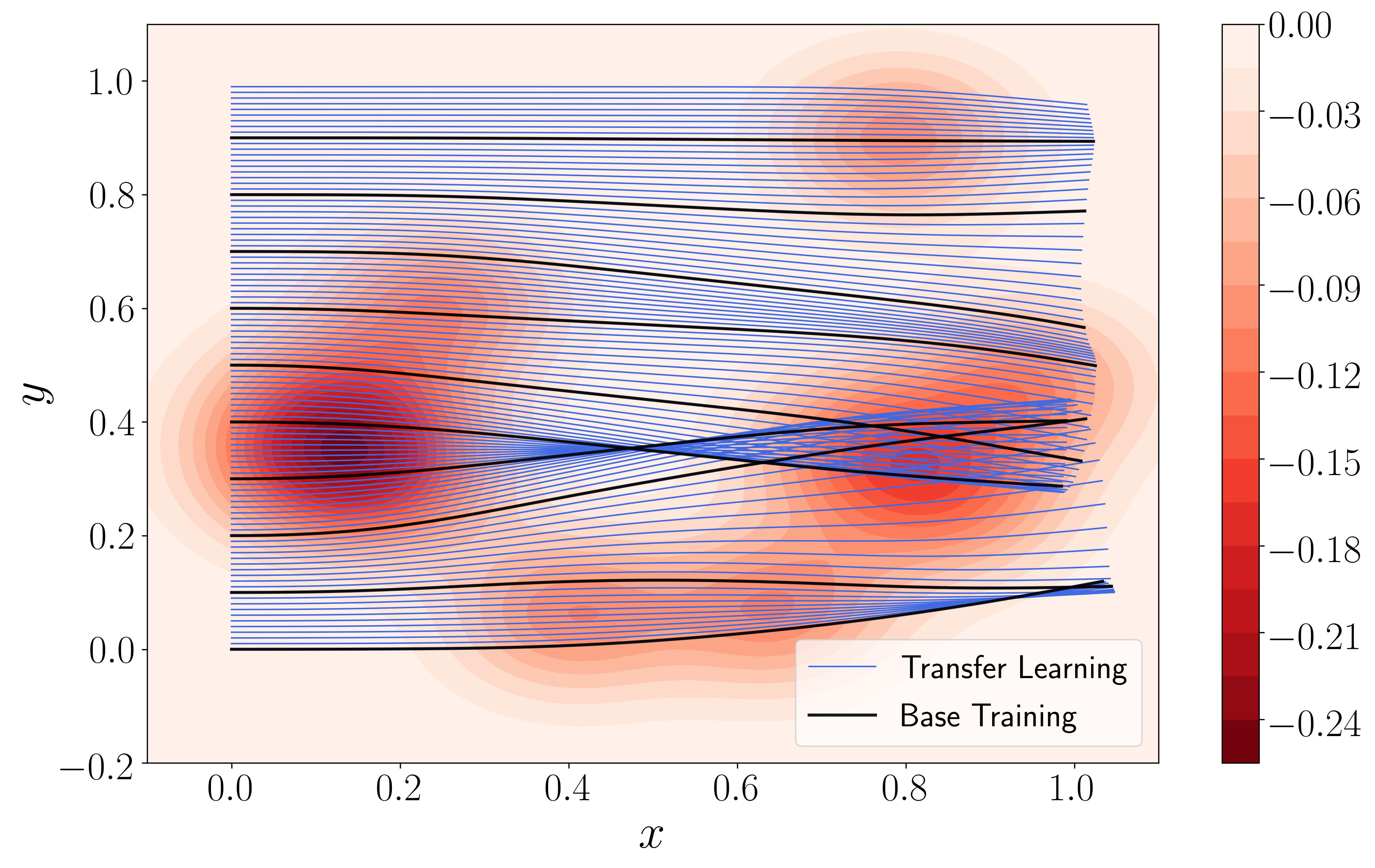}
\caption{Initial Condition Transfer}
\label{fig:dense_plots_ic}
\end{subfigure}
\hfill
\begin{subfigure}[h]{0.5\linewidth}
\includegraphics[width=\linewidth]{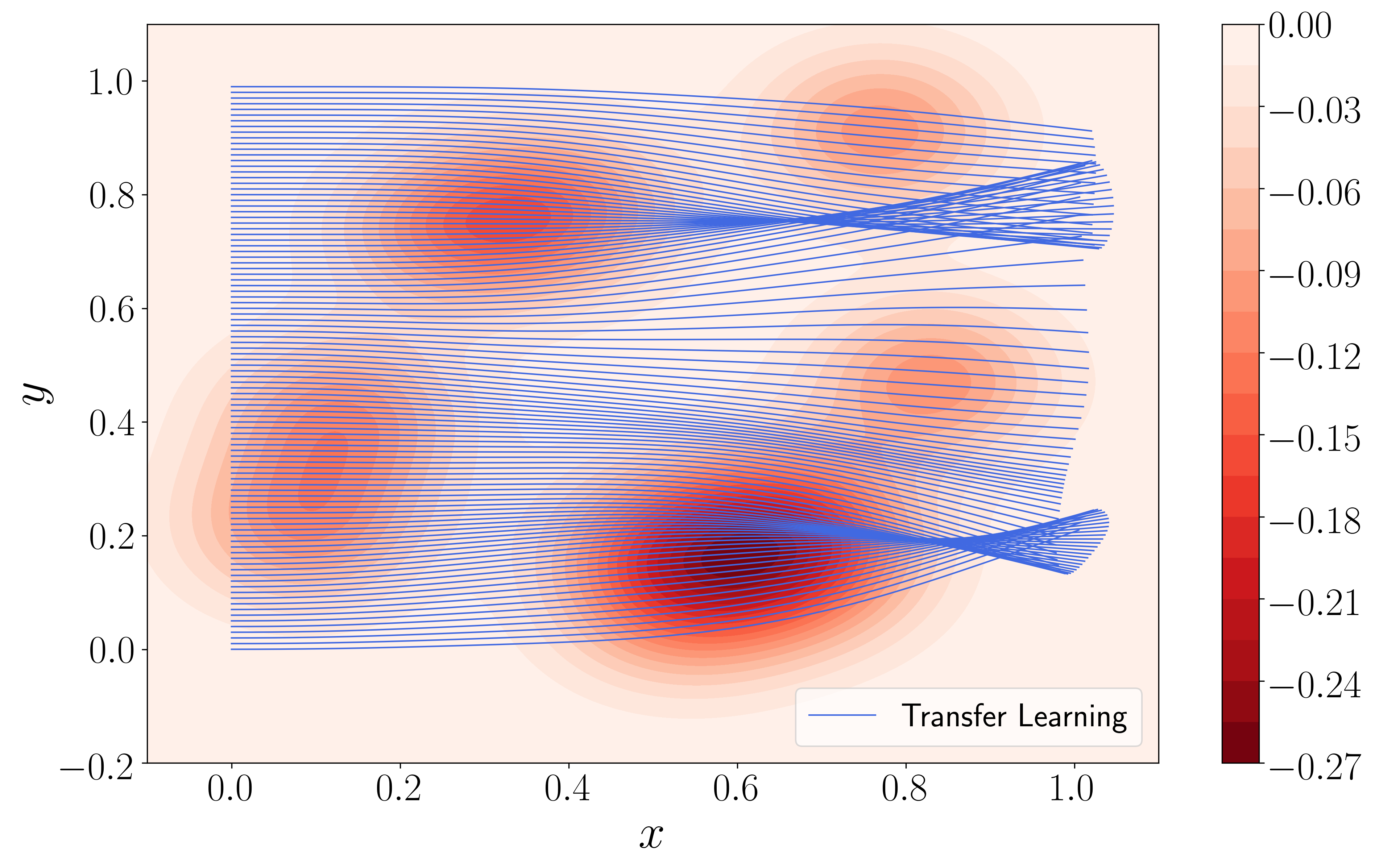}
\caption{Potential Transfer Learning}
\label{fig:dense_plots_potential}
\end{subfigure}%
\caption{Particle trajectories through weak potentials generated by FFNN models. Black lines correspond to the 11 initial conditions used for base training, while blue lines show the solutions obtained for 100 evenly spaced initial conditions via transfer learning after freezing the base. The colorations represent the value of the potential.}
\label{fig:dense_plots}
\end{figure}

\subsection{Details for Hamilton's Equations}

For both the FFNN and DEQGAN models, we use networks with a base consisting of 5 hidden layers and 40 nodes. We then use $L$ linear layers for the heads. Each head is responsible for the solution to one ray, i.e., one initial condition  $\bold{z}_l(0)=(0,y_l(0),1,0)$, where $l=1,\dots,L$, and has four outputs corresponding to $x$, $y$, $p_x$ and $p_y$. For head $l$, we denote the outputs as $u_{l,1}$, $u_{l,2}$, $u_{l,3}$ and $u_{l,4}$. We use the initial value re-parameterization proposed by
\citet{mattheakis2019physical}
\begin{equation}
    \label{eq:ic_reparam}
    \tilde{u}_{l,i}(t)=\left[\bold{z}_l(0)\right]_i+\left(1-e^{-t}\right)u_{l,i}(t), \quad l=1,\dots,L; i=1,\dots,4
\end{equation}

which forces the proposed solution to be exactly $\bold{z}_l(0)$ when $t=0$ and decays this constraint exponentially in $t$.

\subsection{Transfer Learning Results}

Our first transfer learning task, Initial Condition Transfer, allows us to efficiently obtain solutions to the system for many initial conditions. We used multi-head models to train the base networks on $L=11$ initial conditions $y_l(0)=0.0,0.1,\dots,1.0$ and performed single-head transfer learning on $100$ evenly-spaced initial conditions in $[0,1]$ while keeping the potential fixed. All experiments were performed on a Microsoft Surface laptop with Intel i7 CPU. Figure~\ref{fig:dense_plots_ic} shows the ray trajectory solutions corresponding to the initial conditions used for base training (black) and transfer learning (blue) obtained with the FFNN model. These trajectories also illustrate branched flows.

\begin{figure}
\begin{subfigure}[h]{0.45\linewidth}
\includegraphics[width=\linewidth]{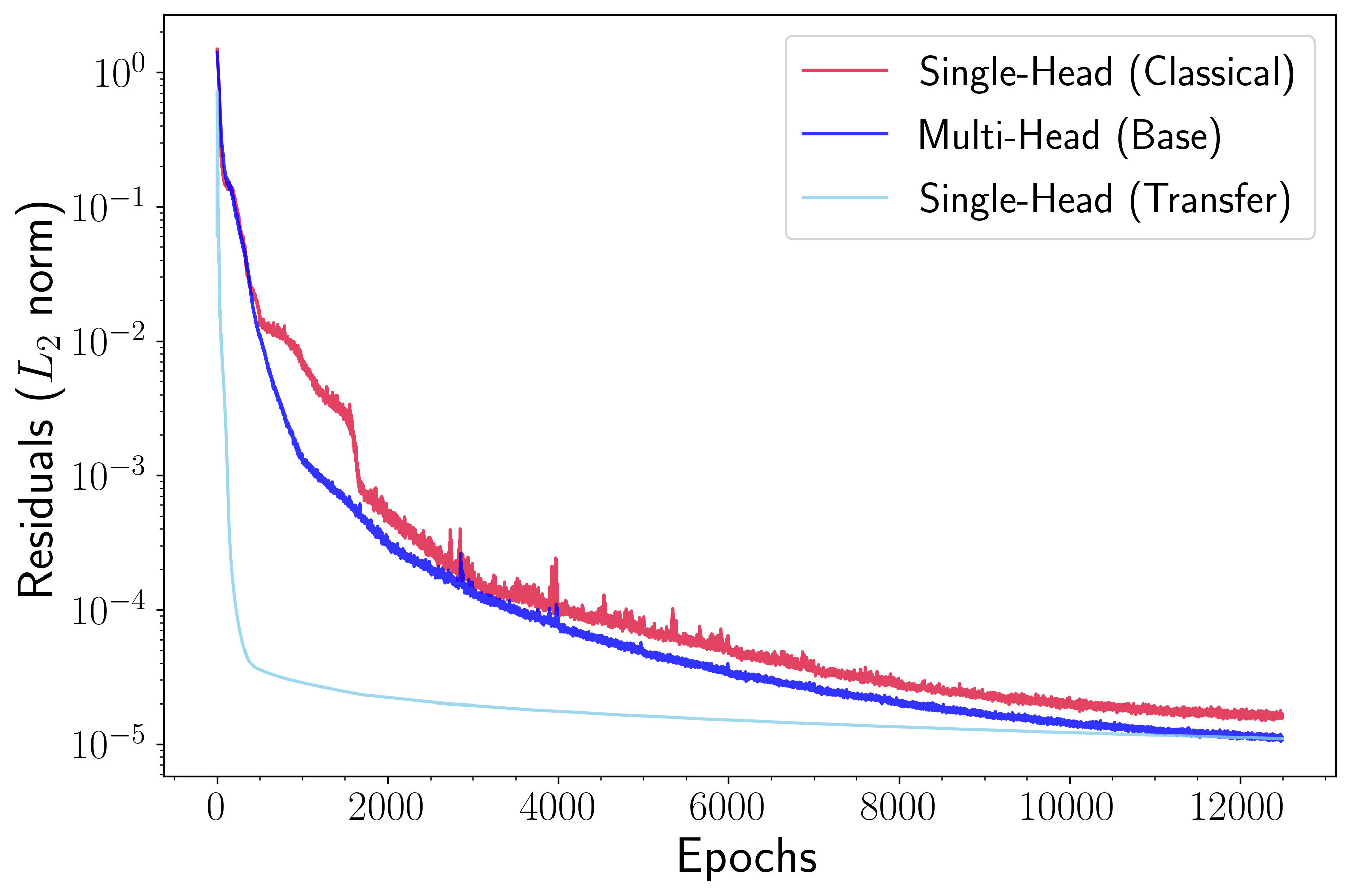}
\caption{FFNN}
\end{subfigure}
\hfill
\begin{subfigure}[h]{0.45\linewidth}
\includegraphics[width=\linewidth]{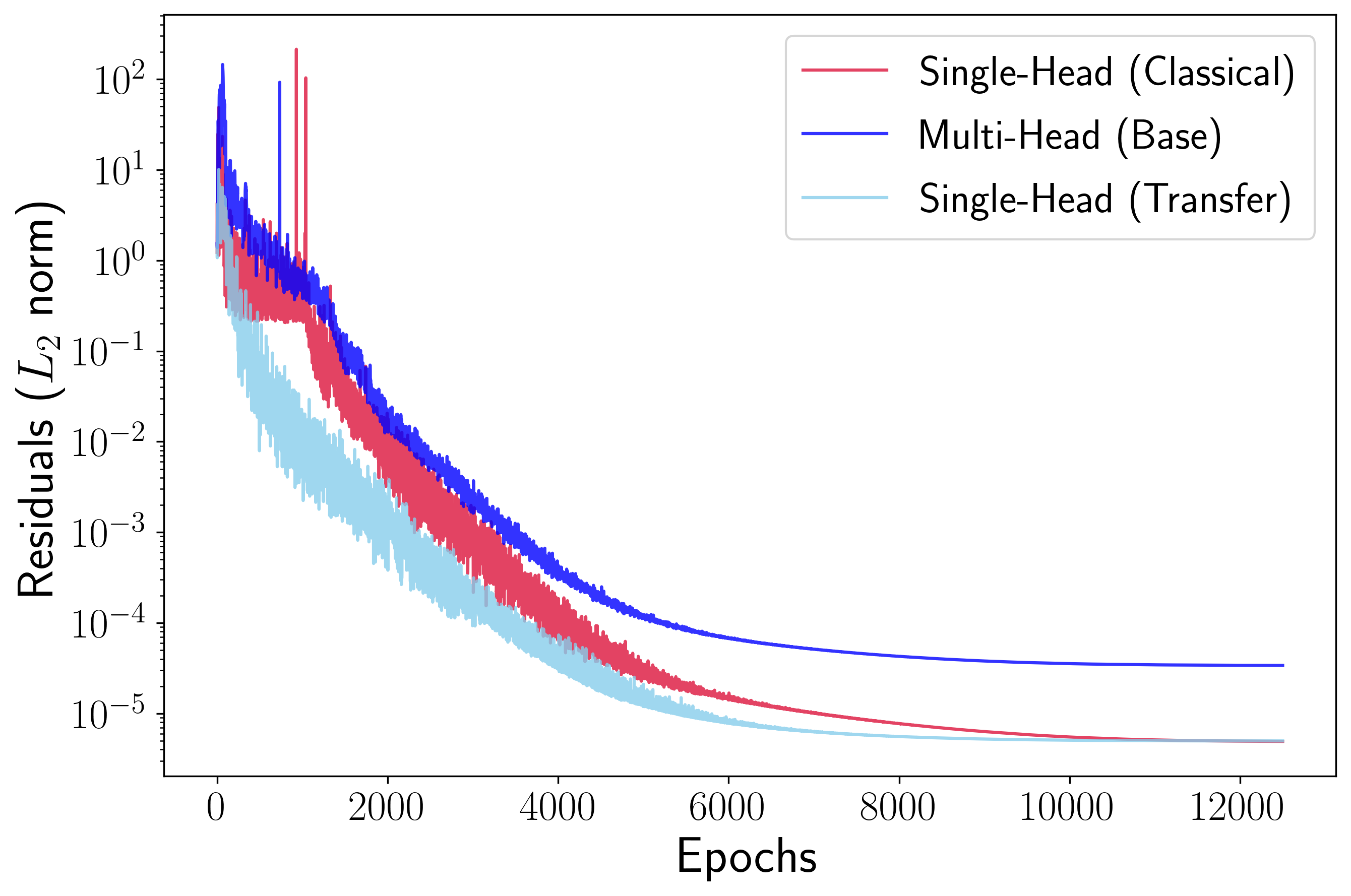}
\caption{DEQGAN}
\end{subfigure}%
\caption{Epochs vs. $L_2$ norm of the equation residuals for single-head (classical), multi-head (base training) and single-head (Initial Condition Transfer) runs using FFNN and DEQGAN models. The multi-head model losses are computed by averaging over the 11 heads.}
\label{fig:classical_base_transfer}
\end{figure}

In Figure~\ref{fig:classical_base_transfer}, we compare the losses achieved by the FFNN and DEQGAN models during base training and transfer learning. We also show the residuals for classical models that do not leverage transfer learning. Notably, we see that single-head models that use transfer learning converge more rapidly than those trained from scratch. Further, Table~\ref{tab:epochs_per_second} shows that each epoch of transfer learning (bold) is also significantly faster. This is to be expected because transfer learning involves only fine-tuning linear heads, rather than training an entire base network.

Our second transfer learning task, Potential Transfer Learning, utilizes the same pre-trained base described above. This task, however, changes not only the initial conditions, but also the potential (Equation~\ref{pot}). More specifically, we constructed a new potential by randomly sampling ten new Gaussian means. To avoid significantly altering the statistical properties of the system, we used the same values of $\sigma$ and $A$. Figure~\ref{fig:dense_plots_potential} visualizes the ray trajectories obtained using this method and suggests that the multi-head models are, indeed, able to learn highly general bases for the system.

\begin{table}[t]
  \caption{Comparison between the computational efficiency (measured in epochs/sec) of training classical single-head PINNs and performing Initial Condition Transfer for FFNN and DEQGAN.}
  \label{sample-table}
  \centering
  \begin{tabular}{llll}
    \toprule
    \multicolumn{4}{c}{Epochs per second}                   \\
    \cmidrule(r){2-4}
         & Single-Head (Classical) & Multi-Head (Base) & Single-Head (Transfer) \\
    \midrule
    FFNN & $35.27$ & $4.13$ & $\mathbf{42.34}$   \\
    DEQGAN & $8.66$ & $1.77$ & $\mathbf{17.09}$  \\ 
    \bottomrule
  \end{tabular}
  \label{tab:epochs_per_second}
\end{table}

\section{Conclusion}

In this paper, we propose a multi-head PINN architecture and a framework for performing transfer learning with non-linear systems of differential equations. In particular, we simulate branched flows with Hamilton's equations and demonstrate that our method significantly reduces the computational cost of obtaining solutions to many initial conditions in comparison to FFNN and GAN-based models trained from scratch, without sacrificing accuracy. Finally, we show that base networks trained using our method can transfer to new initial conditions {and} new potentials at the same time, indicating that our method is able to learn highly general statistical properties of the system.

\section{Broader Impact}

This paper presents techniques that we hope will increase the utility of PINNs in real-world applications. In particular, the transfer learning procedure explored in this work trains a base neural network on different configurations of the system of equations, thereby forcing the network to learn general properties of the solutions and providing possible insights into the underlying physical problem. Beyond computational speedups, we hope that this contributes to broader efforts within the research community to make PINNs more interpretable, and ultimately more widely adopted. We believe that future work focused on the theoretical foundations of PINNs will help cement these models as a third pillar within the study of differential equations, alongside analytical and numerical methods.

\bibliographystyle{apalike2}
\bibliography{references}


\end{document}